\documentclass[letterpaper, 10 pt, conference]{ieeeconf}  % letterpaper or a4paper
\IEEEoverridecommandlockouts                              % For \thanks command
\usepackage{graphicx}  % for pdf, bitmapped graphics files
\usepackage{amsmath}   % assumes amsmath package installed
\usepackage{amssymb}   % assumes amsmath package installed
\usepackage{hyperref}
\usepackage{subcaption}
\newcommand{\vect}[1]{\boldsymbol{#1}}
\title{\LARGE \bf Self-supervised Monocular Multi-robot Relative Localization with Efficient Deep Neural Networks}
\author{Shushuai Li, Christophe De Wagter and Guido C. H. E. de Croon% <-this % stops a space
    \thanks{All authors are with Faculty of Aerospace Engineering,
        Delft University of Technology, 2629 HS Delft, The Netherlands (e-mail: s.li-6@tudelft.nl; c.deWagter@tudelft.nl; g.c.h.e.decroon@tudelft.nl)
    }%
}
\begin{document}
\maketitle
\thispagestyle{empty}
\pagestyle{empty}

\begin{abstract}
Relative localization is an important ability for multiple robots to perform cooperative tasks in GPS-denied environment.
This paper presents a novel autonomous positioning framework for monocular relative localization of multiple tiny flying robots.
This approach does not require any groundtruth data from external systems or manual labelling.
Instead, the proposed framework is able to label real-world images with 3D relative positions between robots based on another onboard relative estimation technology, using ultra-wide band (UWB).
After training in this self-supervised manner, the proposed deep neural network (DNN) can predict relative positions of peer robots by purely using a monocular camera.
This deep learning-based visual relative localization is scalable, distributed and autonomous.
We also built an open-source and light-weight simulation pipeline by using Blender for 3D rendering, which allows synthetic image generation of other robots, and generalized training of the neural network.
The proposed localization framework is tested on two real-world Crazyflie2 quadrotors by running the DNN on the onboard AIdeck (a tiny AI chip and monocular camera).
All results demonstrate the effectiveness of the self-supervised multi-robot localization method.
\end{abstract}

\section{Introduction}
Relative localization is necessary for a robot to interact with peer robots, underlying a wide range of distributed and cooperative tasks, e.g., formation flight \cite{kushleyev2013towards}, cooperative construction \cite{lindsey2011construction}, flocking behaviour \cite{vasarhelyi2018optimized}, etc.
However, most of the multi-robot systems rely on external devices such as the global positioning system (GPS) or motion capture systems, for providing the relative positions between robots.
These systems cannot work in unknown, GPS-denied environments.

Onboard relative localization methods have been recently proposed for achieving fully autonomous operation of multi-robot systems.
Relative estimation based on sound \cite{basiri2014audio} or infra red \cite{roberts20123} is impractical for nano robots as larger sensor arrays need to be mounted.
Relative localization with communication chips is very suitable for tiny robots thanks to their light weight. For example, multiple tiny quadrotors can avoid each other based on the received signal strength (RSS) \cite{mcguire2019minimal}.
More precise ranging from UWB can be implemented on the same tiny robots for more accurate relative estimation \cite{li2020autonomous}.
However, these methods suffer from band-width limitations, leading to poor scalability for a larger number of robots.

\begin{figure}[ht]
    \centering
    \includegraphics[width=0.45\textwidth, trim={0cm 0cm 0cm 0cm}, clip]{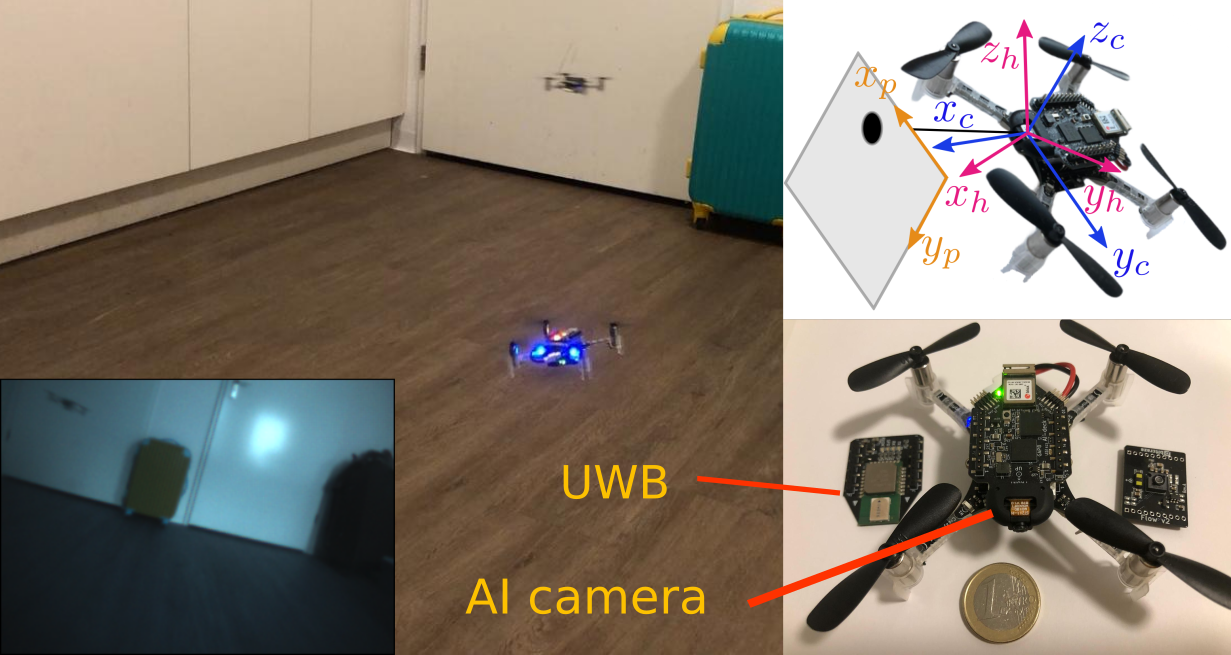}
    \caption{Multiple tiny Crazyflie quadrotors localize peer robot 3D positions with deep neural networks based on self-supervised labels  from an ultra-wide band relative localization method. Top right: different coordinate frames. Bottom right: a tiny quadrotor, the AI camera and the UWB module. Bottom left: onboard captured image.}
    \label{fig:loca_show}
    \vspace*{-0.6cm}
\end{figure}

Vision-based methods are scalable and distributed for multi-robot localization.
These methods can be divided into two main categories: marker-based traditional methods and marker-less learning based detection.
The methods with markers consist of relative localization for multiple micro aerial vehicles (MAVs) with onboard markers \cite{saska2017system}, collaborative localization for a swarm of MAVs relying on salient external features for sparse reconstruction \cite{vemprala2018monocular}, estimation of relative pose between two ground robots by observing a pair of 3D points \cite{rodrigues2019framework}, and two drones tracking same target cooperatively \cite{price2018deep}.
The performance of these methods is easily degraded due to the size of markers (which should be very small for tiny robots) and the marker pose in the image.
Although there is a swarm of quadrotors based on general textures by using visual inertial odometry (VIO) \cite{weinstein2018visual}, the relative pose will drift with time and they must take off from known locations. An improvement for VIO-based relative localization is combining it with the UWB measurement \cite{xu2020decentralized}. However, the VIO part requires considerable computation power, which is impracticable  for tiny flying robots.

Learning-based visual localization is marker-independent and directly robot-oriented. For example, 3D positions of a drone can be estimated by using depth images and deep neural networks \cite{carrio2018drone}.
DeepURL proposes a deep estimation method for relative localization of underwater vehicles based on keypoint prediction and PnP which, however, requires each robot's 3D model information \cite{joshi2020deepurl}.
Another state-of-the-art work uses YOLOv3-tiny to detect the drones by training the network with mask images from a static camera and background subtraction method \cite{schilling2020vision},
in which the deep neural network is too heavy to run on tiny drones.
Besides, its training dataset has a simple background such that the detection will be subject to a larger reality gap caused by common, more cluttered environments, and potentially by motion blur and lighting conditions.

We also review the related references in computer vision and robotic grasping.
Classical pose networks require manual annotations such as
SSD \cite{kehl2017ssd}, PoseCNN \cite{xiang2017posecnn}, and PoseNet \cite{kendall2015posenet}.
Model-based methods can extract more pose information without or with less annotations. For example, EPOS predicts 3D fragment coordinates only with coordinate annotation, and then uses PnP-RANSAC to get 6D object pose \cite{hodan2020epos}. Instead of using coordinate annotation, a deep neural network is designed by detecting 2D projections of robot joints, combined with PnP and model information to estimate the camera-to-robot pose from a single image \cite{lee2020camera}. However, annotations are time-consuming, and 3D model information is not significant for tiny robots.

Extended visual information can facilitate the 6D pose estimation, such as RGB-D images for object pose prediction, based on a model \cite{wada2020morefusion}, or in a self-supervised way by capturing images from different views \cite{deng2020self}.
A pair of images is used for self-supervised depth estimation \cite{hur2020self}. These extended visual sensors are usually heavy, power hungry, and high-cost for tiny robots compared with a monocular camera.

% Flow estimation and self localization, Flownet3d \cite{liu2019flownet3d}, predict relative pose for visual odometry and localization with respect to a path \cite{gridseth2020deepmel}, DFVS integrates optical flow with depth estimates for relative camera pose prediction on Bebop \cite{harish2020dfvs}, Global localization such as OneShot leverages 3D LiDAR point cloud to localize robot itself \cite{ratz2020oneshot}.

This paper proposes a self-supervised framework for autonomous, scalable and low-cost relative localization of multiple tiny robots. The proposed network draws from the object detection research YOLOv3, but is adapted to the multi-robot localization domain. We do not predict bounding boxes. Rather, we predict the 2D pixel position of the robot center and depth from camera to robot. Here we adopt our previous work of UWB-based relative localization as an auxiliary localization method to label the images automatically \cite{li2020autonomous}, to make the training process self-supervised. The \textbf{contributions} of this paper are summarized as: 1) an efficient deep neural network for integrated monocular multi-robot detection and depth prediction; 2) a novel self-supervised system framework for data labelling and network training; 3) a light-weight 3D rendering simulation for multi-robot image generation with arbitrary pose states; 4) the first implementation of deep neural networks into light tiny (33 gram) flying robots for visual relative localization; 5) Public release of all code and dataset to the community \footnote{Code: \url{https://github.com/shushuai3/deepMulti-robot}}.

The remainder of the paper is organized as follows.
Section \ref{sec:loca_system} introduces the system definition and the relative localization problem.
Section \ref{sec:loca_approach} gives the detailed design of the proposed framework and the deep neural network.
Section \ref{sec:loca_sim} shows the 3D multi-robot visual rendering pipeline, and performance of the deep relative localization on synthetic images.
Section \ref{sec:loca_exp} validates the localization efficacy on real-world experiments, including dataset collection, network refining, and deep inference onboard an AI camera.

\section{Preliminaries}
\label{sec:loca_system}
A monocular camera mounted on one robot can observe $n$ peer robots in a single RGB image.
Before exploring 3D relative pose between the camera and peer robots, this section gives the preliminaries of the multi-robot visual system, the auxiliary onboard relative localization, and the problem definition.

\subsection{Multi-robot system}
For clarity, the spatial model of two robots is considered.
As shown in Fig. \ref{fig:loca_show}, we define three coordinates: 1) image coordinate with yellow axes, where $\vect{\xi}_p=[x_p, y_p, 1]^{T}$ denotes pixel positions of peer-robot center in the image with top-left origin;
2) camera coordinate with blue axes, where $\vect{\xi}_c=[x_c, y_c, z_c]^{T}$ represents 3D positions of peer robot with respect to the camera; 3) horizontal coordinate with red axes, which is an inertial frame fixed to the robot with a vertical $z$ axis, while $x$ and $y$ axes point forward and left horizontally.

Camera coordinates can be transformed to image coordinates by the intrinsic matrix $M_{\rm{itr}}$ of the camera. The intrinsic parameters of the AIdeck camera is calibrated with a chessboard, and $R_c$ means the rotation of the camera coordinate, which are shown as follows:
\begin{equation} \label{eq:loca_imageCamera} \setlength{\arraycolsep}{0.5pt} {\medmuskip=0mu
x_c\vect{\xi}_p = M_{\rm{itr}}R_c\vect{\xi}_c = 
\begin{bmatrix}
    183.73 & 0 & 166.90 \\
    0 & 184.12 & 77.51 \\
    0 & 0 & 1
\end{bmatrix}
\begin{bmatrix}
    0 & -1 & 0 \\
    0 & 0  & -1\\
    1 & 0 & 0
\end{bmatrix}\vect{\xi}_c }
\end{equation}

The auxiliary 3D relative estimation is represented in the horizontal coordinate, denoted as $\vect{\xi}_h=[x_h, y_h, z_h]^{T}$, which facilitates real-world 3D multi-robot control. This coordinate can be transformed into camera coordinate by rotation in xy sequence
\begin{equation} \label{eq:loca_horizon2camera} \setlength{\arraycolsep}{1.5pt}
\vect{\xi}_c = R(\phi,\theta)\vect{\xi}_h =
\begin{bmatrix}
    c(\theta) & 0 & s(\theta) \\
    s(\phi)s(\theta) & c(\phi) & -c(\theta)s(\phi) \\
    -c(\phi)s(\theta) & s(\phi) & c(\theta)c(\phi)
\end{bmatrix}
\vect{\xi}_h,
\end{equation}
where $\phi$ and $\theta$ denote roll and pitch attitude along axis $x_c$ and $y_c$ respectively.

\subsection{Onboard auxiliary localization}
This subsection gives a brief review of the UWB-based relative estimation \cite{li2020autonomous}, which is used for generating labels to teach the deep neural network to learn peer-robot positions from monocular images.

\begin{figure*}[ht]
    \centering
    \includegraphics[width=0.95\textwidth]{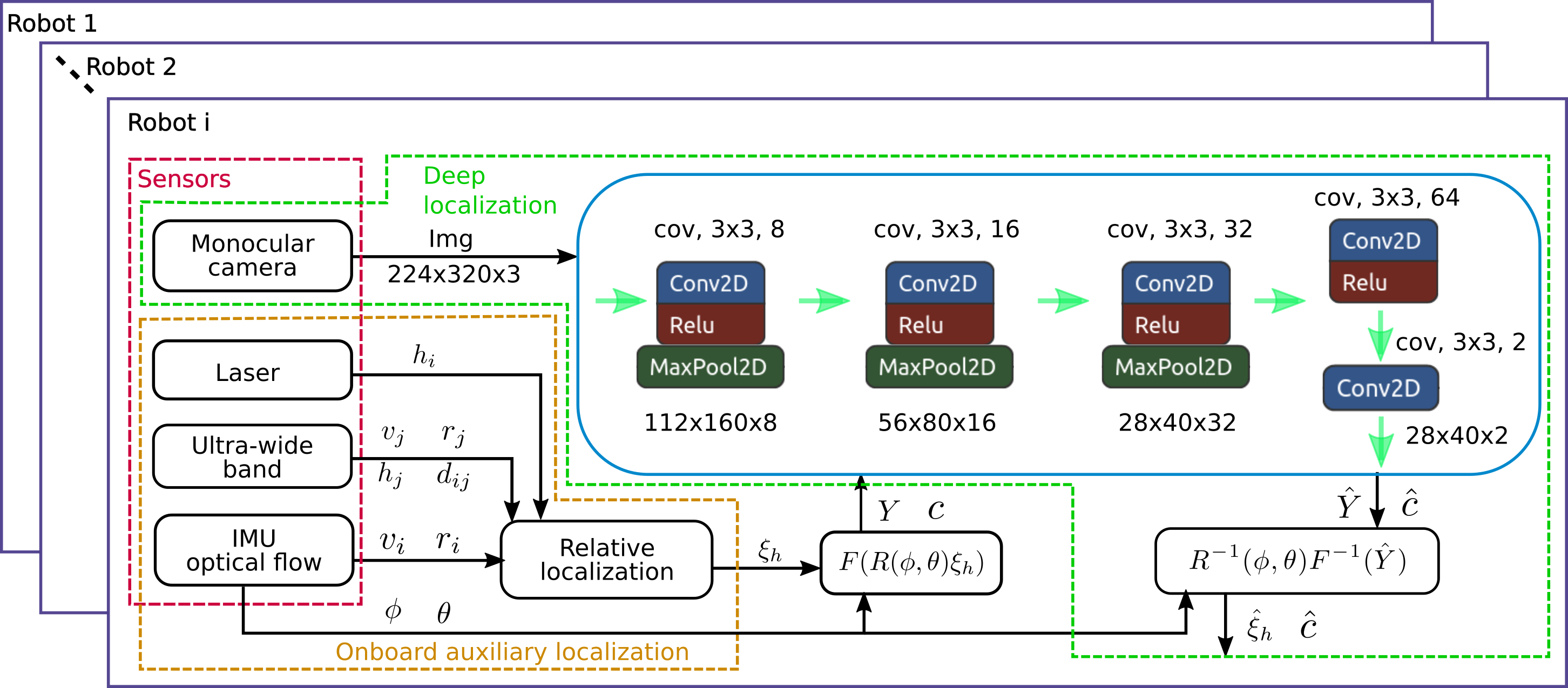}
    \caption{System framework for deep relative localization of tiny flying robots. Red block shows all onboard sensors. Yellow block is the onboard auxiliary localization, which outputs the relative position label. This data is transformed to pixel position and depth for training the neural network as shown in the green block. After training, the deep localization can use purely monocular image to estimate the 3D multi-robot relative pose. In addition, the network architecture consists of convolution layers, Relu activation, and max pooling layers as shown in the blue block. It also shows the kernel size, input and output channels, and the image size for each step.}
    \vspace*{-0.3cm}
    \label{fig:loca_network}
\end{figure*}

As can be seen in Fig. \ref{fig:loca_network}, the yellow box illustrates the auxiliary localization. The ${i}^{\rm{th}}$ robot takes as inputs the peer velocity $v_j$, peer yaw rate $r_j$, peer height $h_j$, self velocity $v_i$, self yaw rate $r_i$, self height $h_i$, and range $d_{ij}$. Afterwards, a Kalman filter is implemented to estimate the relative position $[x_{ij}, y_{ij}, h_{ij}]^{T}$, which is equal to $\vect{\xi_h}$. More details can be found in \cite{li2020autonomous}.

\subsection{Self-supervised localization problem}
\label{sec:loca_problem}
Suppose the multi-robot system is equipped with the aforementioned localization techniques.
Each robot captures the monocular image and obtains the corresponding label $\vect{\xi}_h$ automatically.
Given the image $Im$ with a peer robot in it, the self-supervised deep localization problem is to find a deep neural network $f_n(Im)$ that can predict peer-robot relative positions $\vect{Y}$ which satisfies $\vect{Y}=[x_p, y_p, x_c]^{T}$.
According to \eqref{eq:loca_imageCamera} and \eqref{eq:loca_horizon2camera}, the desired relative positions $\vect{Y}$ can be derived from automatic label $\vect{\xi}_h$ as:

\begin{equation} \label{eq:loca_problem} {\medmuskip=0mu
\begin{aligned}
\vect{Y}&=F(R(\phi,\theta)\vect{\xi}_h)=[\vect{\xi}_p^{[0:2]}; x_c]\\
&=[(M_{\rm{itr}}R_cR(\phi,\theta)\vect{\xi}_h/x_c)^{[0:2]}; x_c]
\end{aligned}
}
\end{equation}
where $\vect{\xi}_p^{[0:2]}$ means the first two rows of the vector $\vect{\xi}_p$.
In addition, multiple robots can appear on one monocular image. Thus, the deep inference function $f_n(Im)$ must be able to predict 3D positions of multiple robots.

\section{Methodology}
\label{sec:loca_approach}

This section describes the detailed approach to self-supervised monocular relative localization. The system framework, network architecture and loss functions are explained, respectively.

\subsection{Network output and self-supervised dataset}
The data flow of the whole system is shown in Fig. \ref{fig:loca_network}.
As can be seen, it starts from the auxiliary localization block, which generates the labeled dataset automatically for self-supervised training.
The dataset consists of monocular images and the corresponding inter-robot positions $\vect{Y}=[x_p, y_p, x_c]^{T}$.
To detect multiple robots, the network output $f_n(Im)$ is designed as a 28x40 grid map with the predicted depth channel $\hat{d}(i,j)$ and the confidence channel $\hat{c}(i,j)$. A higher dimensional grid map leads to more accurate pixel localization, which however increases the ambiguity between detections at neighbor pixels.
The grid labels are created by the following rules:
\begin{equation} \label{eq:loca_labelRules} {\medmuskip=0mu
    \begin{cases} 
c(i,j)=1, d(i,j)=x_c, & \rm{if} (i,j) =(x_p/8,y_p/8)\\
c(i,j)=0, d(i,j)=0,   & \rm{otherwise}
\end{cases} }
\end{equation}
which means a grid that contains a robot has confidence of 1 and the depth of $x_c$ in camera coordinate. Since $x_p\in[0,320)$ and $y_p\in[0,224)$, $(x_p,y_p)/8$ fits with the network output size.

In order to obtain a more generalized network, we pretrain the network on a synthetic dataset. In this dataset, the grid labels can be created automatically by the masks of robots during 3D rendering. The synthetic dataset contains more different backgrounds and arbitrary random attitudes and positions of the drones.

\subsection{Network architecture}
Our deep network is inspired by YOLOv3 network which can detect multiple small objects with bounding boxes \cite{redmon2018yolov3}. We modify the YOLOv3 network to solve the localization problem as described in Section \ref{sec:loca_problem}. The proposed network predicts the pixel position and depth of peer robots as can be seen in the blue block in Fig. \ref{fig:loca_network}, instead of bounding boxes as the original YOLOv3 network. The feature maps and layers of the original network are largely reduced, as we only predict one class of robots. Also, this simplified network fits with the implementation on a resource-limited tiny AI chip, the GAP8 microprocessor, which was first demonstrated for autonomous corridor following in \cite{palossi201964}.

Specifically, the proposed network is an encoder with eight main layers including convolution and max pooling. The activation adopts the Relu function, and there is batch normalization during the training process. No anchors are required as it focuses on the center of the object. The output layers are modified to predict the confidence grid map for localizing the robot center, and the depth grid map.

\subsection{Loss functions}
The loss functions are also different from those in object detection networks. The total loss of the proposed network is composed by two individual items:
\begin{equation}\label{eq:loca_loss_all}
    l = l_{\rm{d}}+l_{\rm{c}}%+l_{\rm{prob}}. l_{\rm{pix}}+
\end{equation}

Depth loss $l_{\rm{d}}$ denotes the mean of square errors between estimated depth $\hat{x}_c$ and real value $x_c$ in grid $(i,j)$, which is represented by
\begin{equation} \label{eq:loca_loss_d}
    l_{\rm{d}} = \rm{mean}(\sum_{i=1}^{N_{yc}}\sum_{j=1}^{N_{xc}}c(i,j)*(\hat{d}_c(i,j)-d(i,j))^2)
\end{equation}
where $c(i,j)$ is the real confidence of whether there exists an robot center in grid $(i,j)$ defined in \eqref{eq:loca_labelRules}. $N_{yc}=28$ and $N_{xc}=40$ are the sizes of output grid maps in the proposed network.

Confidence loss item $l_{\rm{c}}$ is designed by softmax cross entropy function \cite{redmon2018yolov3}, and the formula is
\begin{equation} \label{Eq:loca_loss_conf}
    l_{\rm{c}}=\rm{mean}(\sum_{i=1}^{N_{yc}}\sum_{j=1}^{N_{xc}}(c-\hat{c})^2[-c*\mathrm{log}(\hat{c})-(1-c)*\mathrm{log}(1-\hat{c})])
\end{equation}
where $\hat{c}(i,j)$ and $c(i,j)$ are the predicted and real confidence in grid $(i,j)$ defined in \eqref{eq:loca_labelRules}.

An optional loss item is to distinguish multiple classes of robots by using the one-hot vectors.
$l_{\rm{prob}}$ demonstrates the class probability error, written as
\begin{equation} \label{Eq:loca_loss_prob}
    l_{\rm{p}} =\rm{mean}(\sum_{i=1}^{N_{yc}}\sum_{j=1}^{N_{xc}}c(i,j)*[-p*\rm{log}(\hat{p})-(1-p)*\rm{log}(1-\hat{p})])
\end{equation}
where $\hat{p}(i,j)$ and $p(i,j)$ are predicted and real probability of different classes in grid $(i,j)$. This item has been tested to be effective for robot classification, but not included in this paper for a better quantization of the network to run in the microprocessor.

\subsection{Training and post processing}
The proposed deep relative localization network is trained from scratch in the environments of Anaconda and Tensorflow2. There are 25 epochs for total training including 2 warm epochs to reduce the primacy effect and avoid the early over-fitting. Given 800 training images and a batch size of five images, each epoch has 160 steps. The learning rate changes adaptively with Adam method. The training can be either run on GPU or CPU as the network size is very small.

After training the network on synthetic images, the network is refined on the real-world dataset (192 training images) captured on two Crazyflies.
The refined model file is quantized into int8 format by the GAP8 tool chains.
Finally, it is compiled to a C function that can run on the GAP8 microprocessor.

Based on the prediction $\hat{c}(i,j)$ and $\hat{d}(i,j)$ from the network $f_n(Im)$, the final relative position $\vect{Y}$ can be calculated by
\begin{equation} \label{eq:loca_pos_pixel}
    [\hat{x}_p, \hat{y}_p, \hat{x}_c]^{\rm{T}} = \{[8*i,8*j,\hat{d}(i,j)]^{\rm{T}})|\hat{c}(i,j)>T_c\}.
\end{equation}
where $T_c$ denotes a proper confidence threshold, which was empirically selected as 0.33 for synthetic dataset and 0.23 for real-world tests.

\section{Simulation}
\label{sec:loca_sim}
This section shows the developed simulation pipeline for synthetic image generation, training and testing results of the network on the synthetic dataset. This simulated environment is not necessary for an onboard deep neural network, but it can generate a flexible and rich multi-robot visual dataset easily for preliminary validation of the DNN.
Refining the network on the simulated dataset saves training time and promises more generality of the neural network.

\subsection{Simulated pipeline for 3D multi-robot rendering}
\begin{figure}[h]
    \centering
    \includegraphics[width=0.48\textwidth]{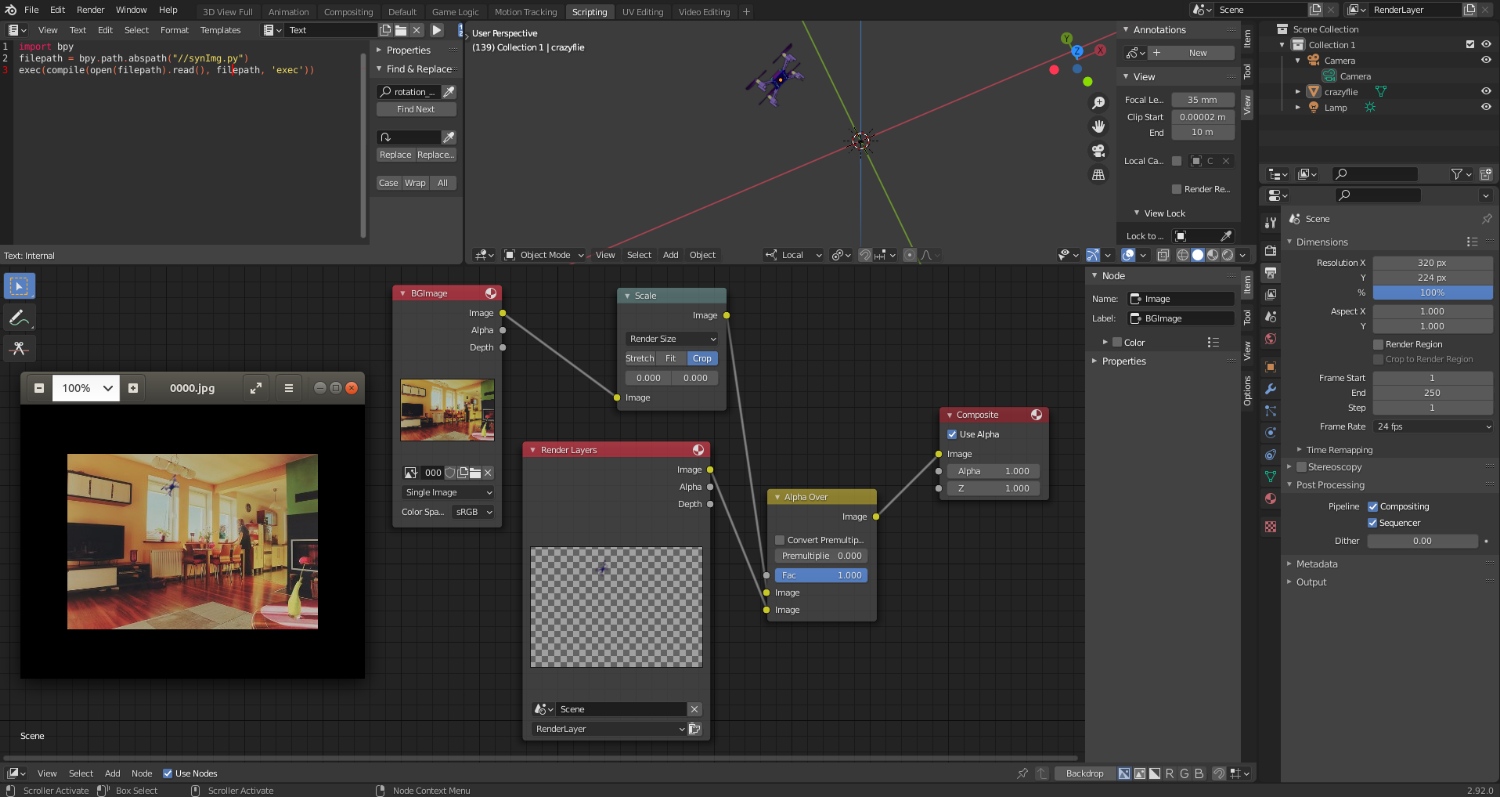}
    \caption{The developed 3D rendering simulation environment. It can render multiple 3D robots in any background images. The attitudes and positions of robots and camera can be set in python code.}
    \label{fig:loca_sim_platform}
\end{figure}
\begin{figure*}[ht]
    \centering
    \includegraphics[width=0.98\textwidth]{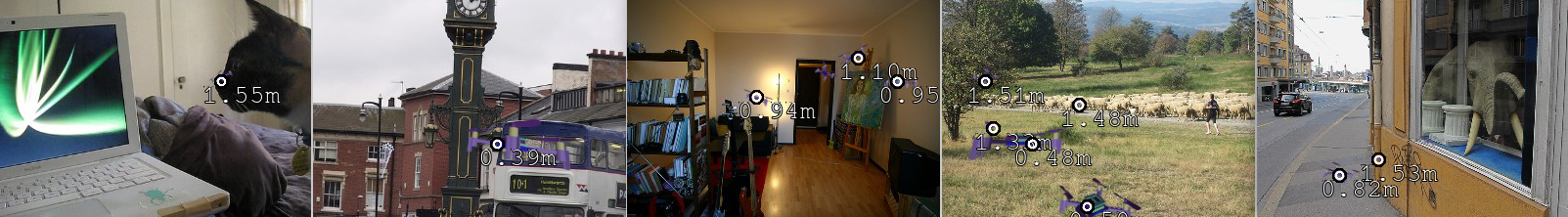}
    \caption{Testing results of multi-robot localization on synthetic images. The white circle with an outer black circle represents the predicted robot position in the image. Different robot attitude and position with respect to the camera are demonstrated in these figures, as well as multiple robots in indoor and outdoor environments.}
    \vspace*{-0.4cm}
    \label{fig:loca_test_syn}
\end{figure*}

We built a Blender-based 3D rendering environment as shown in Fig. \ref{fig:loca_sim_platform} for generating synthetic images in a multi-robot domain. The whole pipeline is light and can render 3D images with a random number of robots in the images with random attitude and positions. The attitude of the camera can be also set arbitrarily. The background images are from the COCO dataset, specifically the 2017-Val-images set including 5000 images. Annotation of robot positions and depth to the camera can be obtained from the prior known groundtruth in Blender. The rendered images can be seen in Fig. \ref{fig:loca_test_syn}. This tool is light-weight, open-source and easily modified for generating multi-robot images for other applications.

\subsection{Training on synthetic dataset}
\begin{figure}[h]
    \centering
    \begin{subfigure}[b]{0.23\textwidth}
        \centering
        \includegraphics[width=\textwidth]{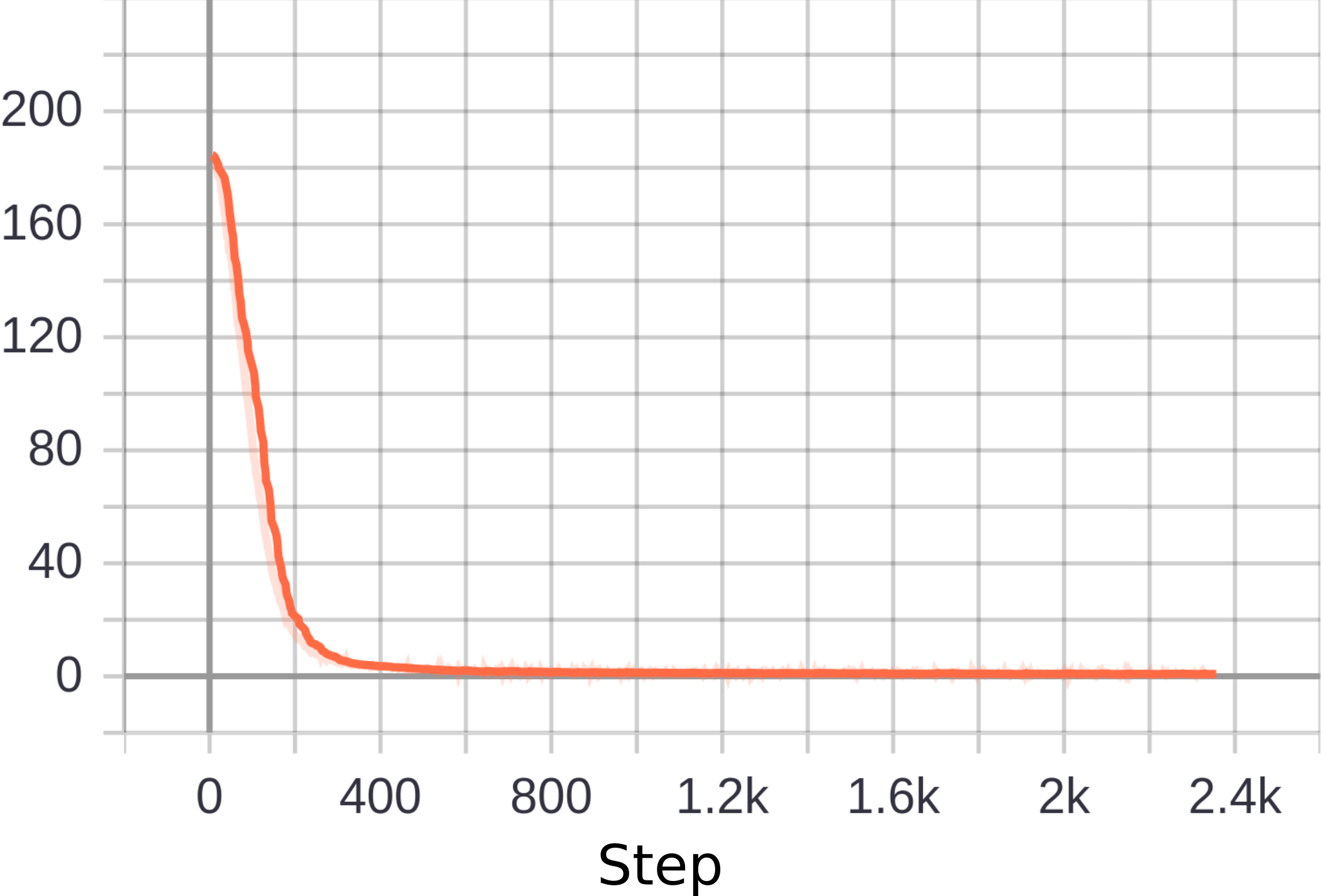}
        \caption{Confidence loss}
        \label{fig:loca_sim_cLoss}
    \end{subfigure}
    \hfill
    \begin{subfigure}[b]{0.23\textwidth}
        \centering
        \includegraphics[width=\textwidth]{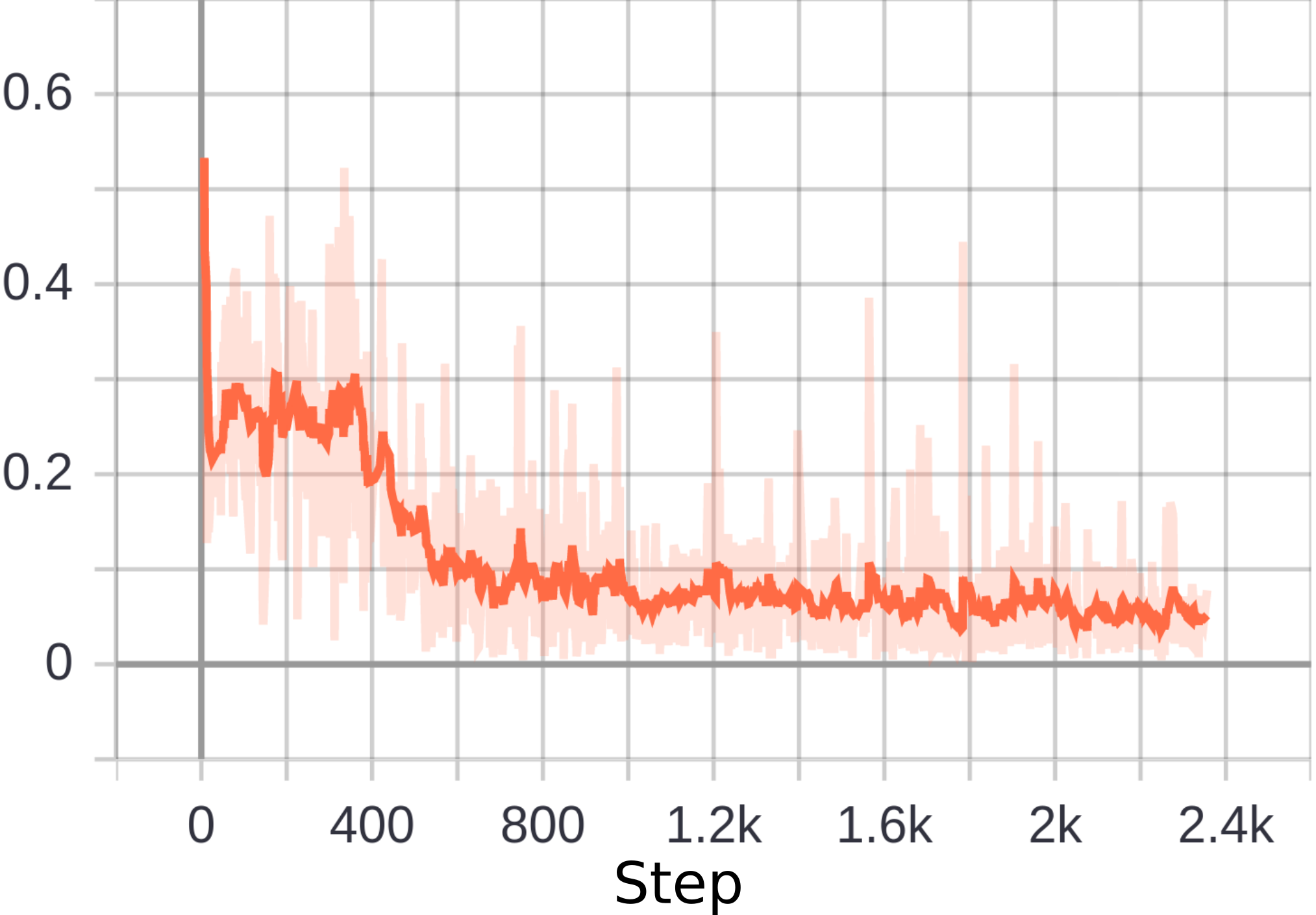}
        \caption{Depth loss (meter)}
        \label{fig:loca_sim_dLoss}
    \end{subfigure}
       \caption{Loss changes with training steps on synthetic dataset. The network is trained from scratch on a training dataset with 800 images. Two individual loss items are shown in this figure.}
       \label{fig:loca_sim_loss}
       \vspace*{-0.2cm}
\end{figure}

From Fig. \ref{fig:loca_sim_loss}, we can see all loss items decrease as the training steps increase. They are stable at about 1000 steps. Specifically, confidence error tends to converge to zero which indicates that the deep neural networks encode the robot pixel position in the output grid maps. The stable depth error is about 0.05m, which is not approximating zero probably because different robot attitudes lead to slight depth changes. However, it is accurate enough for robot position estimation.

\subsection{Testing results on synthetic images}
The prediction errors are shown in this subsection. All testing results come from network prediction on the test dataset which contains 200 new images with different backgrounds and different poses of peer robots.

The preliminary testing results are shown in Fig \ref{fig:loca_test_syn}. From these figures, we can see the proposed network can detect different sizes of robots in outdoor and indoor environments. In addition, thanks to the underlying (modified) YOLOv3 framework, the network is capable to detect multiple robots in one image at same time, even though it is only trained on images with single robot. There are a false-positive detection on the fourth image and a true-negative detection on the third image, potentially due to similar background and less features. These outliers can be rejected by filters in real-time sequences of images. Note that these five test images are selected randomly. Hence, the deep network has similar effectiveness on the other testing images.

\begin{figure}[ht]
    \centering
    \includegraphics[width=0.48\textwidth, trim={1cm 1cm 1cm 1cm}, clip]{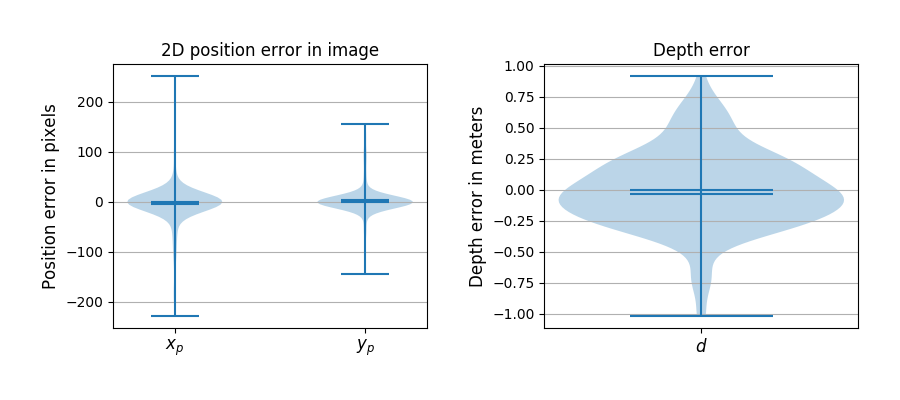}
    \caption{2D robot position error (left figure) in image and depth error (right figure) between prediction and groundtruth are shown. This distribution is based on 200 testing images with single robot in each image.}
    \vspace*{-0.2cm}
    \label{fig:loca_test_error}
\end{figure}

For explicit demonstration, the statistical 3D estimation error is depicted by Fig. \ref{fig:loca_test_error}. From the left figure, we can see the robot localization prediction in image has a zero average pixel error. Most position error is within 20 pixels and with few outliers caused by a few false detections. The right figure demonstrates the error distribution of depth predictions, where mean and medium values are approximating zero.

Therefore, compared to other state-of-the-art research in relative localization with deep learning, our proposed network is much smaller and thus efficient for both training (20 minutes on i7 CPU) and testing. Besides, the drone depth is predicted simultaneously with the drone detection, while other references require the prior-known drone size to estimate its depth.

\section{Experiments}
\label{sec:loca_exp}
This section presents practical experiments: flight for real-world dataset collection, refining of the neural network on the real dataset, porting the Tensorflow-based network to AIdeck microprocessor, and onboard deep visual localization.

\subsection{Hardware}
The multi-robot platform is composed by two Crazyflie2 quadrotors, which are tiny flying robots with only 33 grams and pocket size (12.5 cm in diameter). Three decks are required for our work. The first one is AI deck \cite{palossi201964}, which is composed by a GAP8 RISC-V processor, a Himax HM01B0 RGB camera, 512 Mbit HyperFlash for storing dataset, and an ESP32 wifi module for remote streaming. Another deck is the optical flow deck for estimating robot velocity and altitude. The last one is the loco deck with a UWB module for inter-robot ranging measurements and auxiliary localization.

The last two decks are used for generating relative position labels automatically, while only the first deck is used for deep visual multi-robot localization.

\subsection{Real-world dataset acquisition}
Due to the reality gap between synthetic and real images, a real-world dataset with 240 images is collected for refining the network.
During data collection, one drone flies with randomly velocities by a remote controller, while another drone flies with random velocities but in the view of the camera on the first drone. The example dataset can be found in following sections.

To get dataset among two flying tiny drones, a specific procedure is designed: a quadrotor sends its 2D attitude (roll and pitch) and 3D auxiliary relative position to the GAP8 chip, which combines these variables with the monocular image data and store them in HyperFlash memory. Afterwards, the GAP8 sends the data from flash memory to a computer via a Olimex ARM-JTAG-20-10 debugger.

\subsection{Refining and testing on real-world dataset}

The training process on the real-world dataset is the same as that in Section \ref{sec:loca_sim}. The loss changes during refining process are shown in the following figure.

\begin{figure}[h]
    \centering
    \begin{subfigure}[b]{0.23\textwidth}
        \centering
        \includegraphics[width=\textwidth]{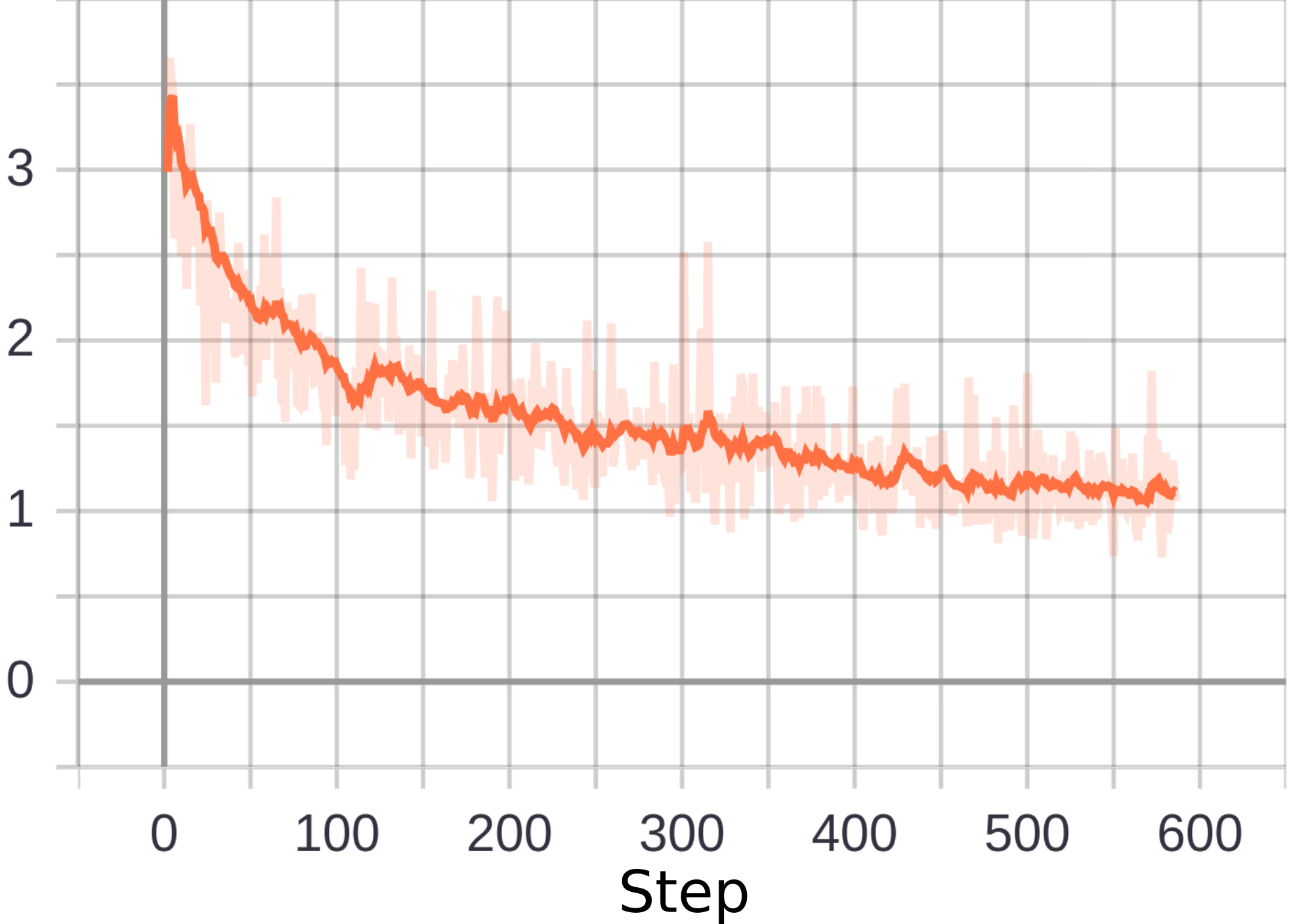}
        \caption{Confidence loss}
        \label{fig:loca_real_cLoss}
    \end{subfigure}
    \hfill
    \begin{subfigure}[b]{0.23\textwidth}
        \centering
        \includegraphics[width=\textwidth]{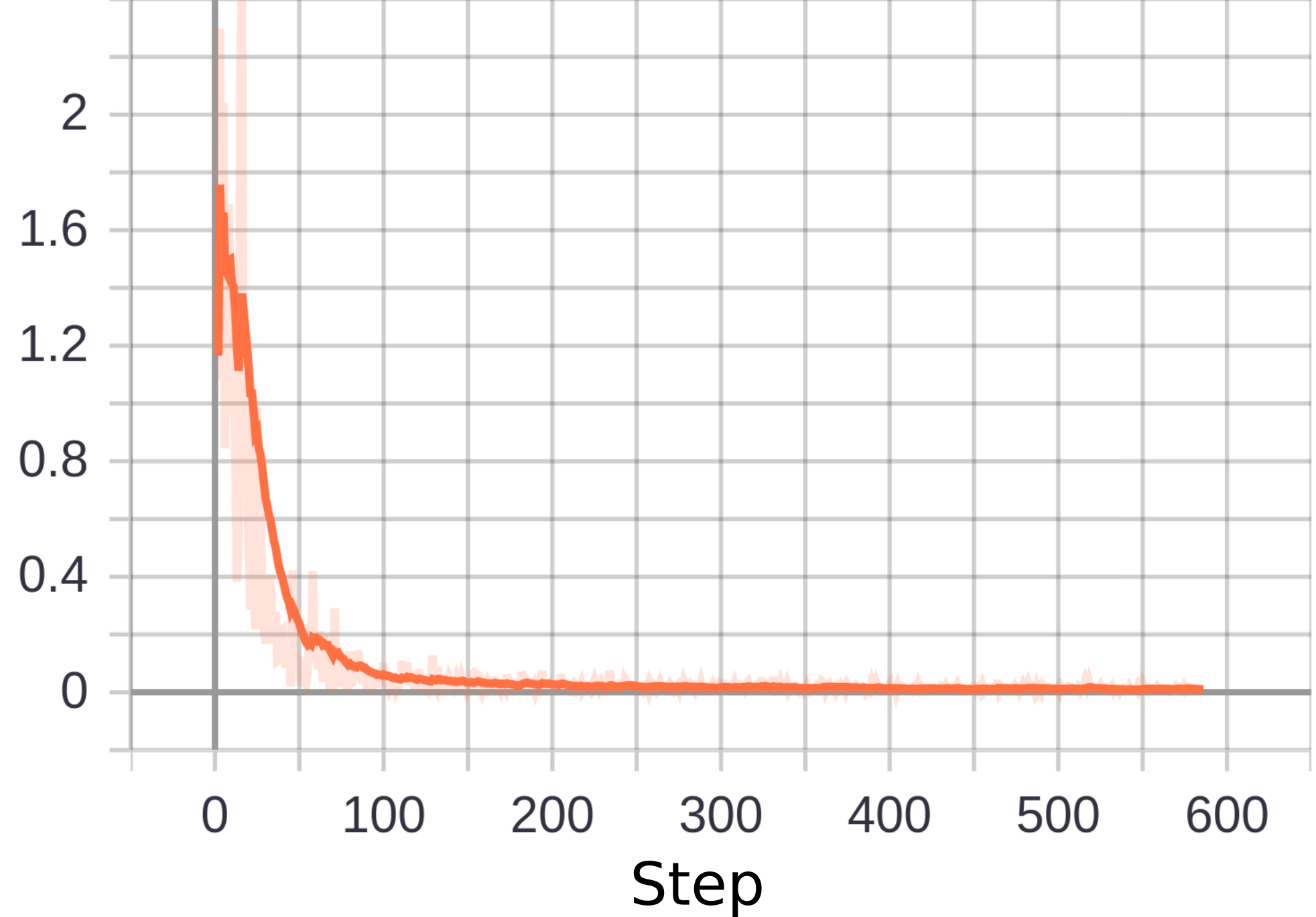}
        \caption{Depth loss (meter)}
        \label{fig:loca_real_dLoss}
    \end{subfigure}
       \caption{Loss changes of refining the neural network on real-world dataset. The initial weights are those trained by the synthetic images in last section. All training configurations remain the same during refining.}
       \label{fig:loca_real_loss}
       \vspace*{-0.5cm}
\end{figure}

From Fig. \ref{fig:loca_real_loss}, we can see all loss items drop down again within 600 steps. The confidence loss decreases a bit slowly than that of depth loss, because the appearance changes largely while drone sizes are easily to learn. The real-world dataset is divided into 192 training images and 48 testing images with self-supervised labels of relative position, all obtained from two randomly flying Crazyflies, without relying on any external systems such as motion capture system or GPS.

\begin{figure}[ht]
    \centering
    \includegraphics[width=0.48\textwidth]{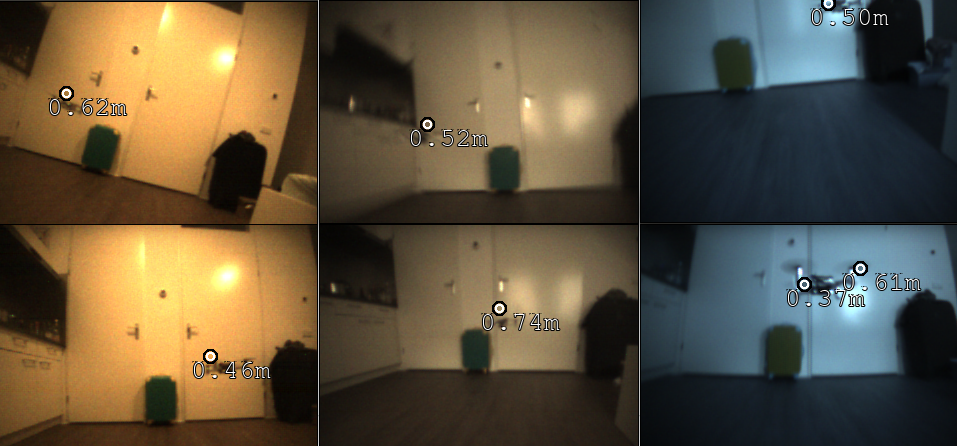}
    \caption{Testing results of the refined network on randomly selected testing images. The white circle in images shows the predicted pixel position of peer robot center. The value on the image means the depth of the robot from the camera. All testing images are captured onboard with different flying attitude and velocity of both quadrotors.}
    \label{fig:loca_test_real}
    \vspace*{-0.2cm}
\end{figure}

Fig. \ref{fig:loca_test_real} shows the localization performance of the refined neural network. Both training and testing images have three light conditions (evening, afternoon and morning, respectively). These testing results indicate that the refined network can localize the real-world flying robots with high accuracy due to the previous training on a more generalized synthetic dataset. It also works even with large motion blur, partial occlusion, and different positions and attitudes of the robots.

\begin{figure}[ht]
    \centering
    \includegraphics[width=0.48\textwidth, trim={1cm 1cm 1cm 1cm}, clip]{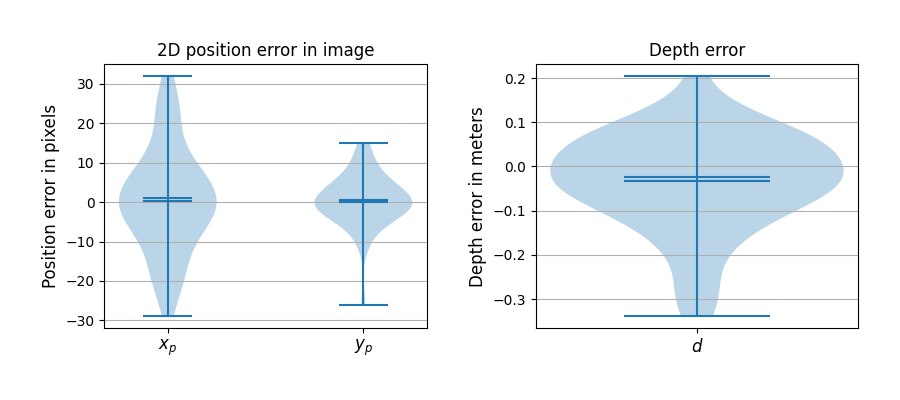}
    \caption{Left: the 2D pixel position error of robot center between deep inference and UWB localization. Right: the depth error distribution in camera coordinate. These results are based on testing dataset with 48 new real-world images with size of 224x320.}
    \label{fig:loca_test_error_real}
    \vspace*{-0.3cm}
\end{figure}

Fig. \ref{fig:loca_test_error_real} demonstrates the statistical prediction errors in image coordinate, by comparing the deep visual localization and auxiliary UWB localization. The refined network has even smaller localization and depth prediction errors than on the synthetic dataset.

% In addition, both figures show a small bias in position prediction, which may come from extrinsic parameters between body coordinate and camera coordinate. This can be compensated by further calibration of the whole system or estimating the bias. Overall, the 3D relative positions are accurate for multi-robot control.

\subsection{Onboard deep relative localization with AIdeck}
This subsection shows the implementation of the deep localization network into a low-cost and ultra low-power AI chip. After quantization from float to int8, the network can run on the AIdeck microprocessor to predict peer-robot position onboard a tiny flying robot.

\begin{figure}[ht]
    \centering
    \includegraphics[width=0.48\textwidth]{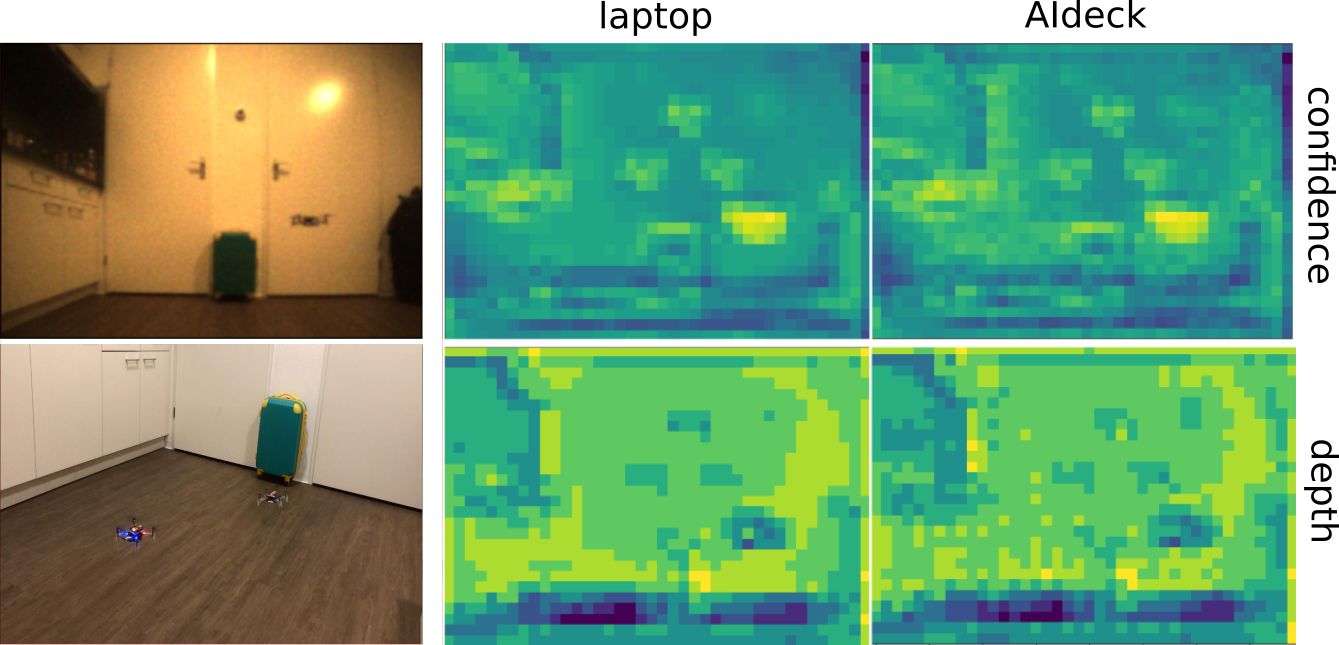}
    \caption{Comparison of the deep visual relative localization between running on the laptop and the onboard AI chip. Top left: an example captured image during flight; bottom left: the experimental flight of two quadrotors; top middle: confidence channel predicted on laptop; bottom middle: depth channel predicted on laptop; top right: confidence channel predicted on AIdeck; bottom right: depth channel predicted on AIdeck.}
    \label{fig:loca_laptopVSaideck}
    \vspace*{-0.2cm}
\end{figure}

Fig. \ref{fig:loca_laptopVSaideck} compares the deep inference results on both laptop (the middle column) and the edge AI chip (the right column) for the same flight image (top left).
The network outputs on PC multiply the quantization scale calculated by the GAP8 tools.
Two right images on first row shows the confidence prediction, where both laptop and AIdeck have similar inference of drone localization in the image. Though they have slight differences in depth prediction as seen in the right two images on second row, the depth values with respect to the robot center area are similar.

\section{Conclusions}
This paper proposes a novel multi-robot relative localization framework which are self-supervised, autonomous, low-cost, and scalable.
Both simulation and experimental results indicate that the proposed deep network can predict peer-robot relative positions with monocular images, without using any external positioning system.
In addition, this paper solves a challenging problem, i.e., implementation of deep neural network into a low-cost tiny AI chip that enables visual multi-robot localization of tiny flying robots.

Future work could include real-world dataset collection in more scenes, and control of a large number of flying robots with the proposed localization method.

%\addtolength{\textheight}{-12cm}  % This command serves to balance the column lengths

%\section*{APPENDIX}

%\section*{ACKNOWLEDGMENT}
%acknowledgments.

\bibliographystyle{IEEEtran}
\bibliography{references.bib}

\end{document}